\RequirePackage{fix-cm}
\PassOptionsToPackage{table}{xcolor}
\documentclass[11pt,letterpaper,shortlabels]{berkeley}
\usepackage[authoryear,round]{natbib}
\usepackage{mathtools,dsfont}
\usepackage{algorithm,algorithmic}
\usepackage{multirow,makecell,array,tabularx}
\usepackage{wrapfig}
\usepackage{xspace}
\usepackage{adjustbox}
\usepackage{tcolorbox}
\ifdefined{lmmono10-regular.otf}\fi 

\DeclareMathSizes{11}{11}{8}{6}

\newcolumntype{Y}{>{\centering\arraybackslash}X}

\definecolor{NDblue}{RGB}{23,50,77}
\definecolor{NDgold}{RGB}{197,139,34}
\definecolor{TableHeader}{RGB}{23,50,77}
\definecolor{TableSubHeader}{RGB}{42,127,158}
\definecolor{TableRowA}{RGB}{247,249,250}
\definecolor{TableRowB}{RGB}{235,242,246}
\definecolor{TableHighlight}{RGB}{224,237,243}
\definecolor{oursbg}{RGB}{252,233,240}
\definecolor{upgreen}{RGB}{0,140,70}
\definecolor{simred}{HTML}{F9D9D7}
\definecolor{oursgreen}{HTML}{D4EDD8}
\definecolor{evidyellow}{HTML}{FFF0A0}

\newcommand{\boldpara}[1]{\par\noindent\textbf{#1}\space}
\newcommand{\rx}[1]{#1$\times$}
\newcommand{\inc}[2][upgreen]{\rlap{\smash{\textsubscript{\textcolor{#1}{\scriptsize (+#2\%)}}}}}
\newcommand{\hlc}[2]{{\setlength{\fboxsep}{0pt}\colorbox{#1}{\strut\,#2\,}}}

\newtcolorbox{promptbox}[2][\small]{
    title={#2},
    colback=TableRowA,
    colframe=NDblue,
    fonttitle=\bfseries\color{white},
    coltitle=white,
    colbacktitle=NDblue,
    boxrule=0.8pt,
    arc=3mm,
    width=\textwidth,
    fontupper=#1
}
\newcommand{\ph}[1]{\texttt{\{#1\}}}                          
\newcommand{\psec}[1]{\par\smallskip\noindent\textbf{#1}\par} 
\newenvironment{pitem}{\begin{itemize}[leftmargin=1.2em,itemsep=1pt,topsep=1pt,parsep=0pt]}{\end{itemize}}

\newcommand{\papertitle}{Twists and Turns: A Graph-Based Benchmark and Framework for Long-Term Memory in Exploration Tasks}
\hypersetup{colorlinks=true,linkcolor=NDblue,citecolor=NDgold,urlcolor=NDblue,
  pdftitle={Twists and Turns: A Graph-Based Benchmark and Framework for Long-Term Memory in Exploration Tasks},
  pdfauthor={Chongyuan Dai}}
\title{\papertitle}
\correspondingauthor{Xiuying Chen (\href{mailto:xiuying.chen@mbzuai.ac.ae}{xiuying.chen@mbzuai.ac.ae})}

\renewcommand{\titlefont}{\color{BerkeleyBlue}\normalfont\bfseries
  \fontsize{17.5}{20}\selectfont}
\makeatletter
\newcommand{\finishwrap}{\par
  \ifx\parshape\WF@fudgeparshape
    \ifnum\c@WF@wrappedlines>2
      \vskip\dimexpr\c@WF@wrappedlines\baselineskip-2\baselineskip\relax
    \fi
    \WFclear
  \fi}
\makeatother

\newcommand{\paperinstitutionsymbol}[1]{%
  \textcolor{NDblue}{\ensuremath{\ifcase#1\relax
    \or\spadesuit\or\heartsuit\or\clubsuit\or\diamondsuit\fi}}}
\newcommand{\papercontactmark}{\textcolor{NDgold}{*}}
\newcommand{\paperauthor}[2]{%
  \mbox{#2\textsuperscript{\paperinstitutionsymbol{#1}}}}
\newcommand{\paperaffil}[2]{%
  \mbox{\textsuperscript{\paperinstitutionsymbol{#1}}\,#2}}
\newcommand{\papertitlerule}{%
  {\color{NDgold}\rule{26pt}{0.6pt}}%
  {\color{NDblue!28}\rule{\dimexpr\linewidth-26pt\relax}{0.6pt}}}
\makeatletter
\renewcommand{\@author}{%
  {\normalfont\fontsize{11}{14}\selectfont
   \paperauthor{1}{Chongyuan Dai}\hspace{0.7em}%
   \paperauthor{2}{Yaling Shen}\hspace{0.7em}%
   \paperauthor{1}{Zirui Song}\hspace{0.7em}%
   \paperauthor{1}{Junfeng Liao}\hspace{0.7em}%
   \paperauthor{3}{Jinpeng Hu}\hspace{0.7em}%
   \paperauthor{1}{Xiuying Chen}\textsuperscript{\papercontactmark}\hspace{0.7em}%
   \par}

  \vspace{6pt}%
  {\normalfont\fontsize{10.5}{14}\selectfont
   \paperaffil{1}{Mohamed bin Zayed University of Artificial Intelligence (MBZUAI)}\\%
   \paperaffil{2}{Monash University}\hspace{1.2em}%
   \paperaffil{3}{Hefei University of Technology}\par%
   }}

\renewcommand{\maketitle}{%
  \begingroup
  \setlength{\parindent}{0pt}%
  \setlength{\parskip}{0pt}%
  \centering
  \vspace*{5pt}%
  {\titlefont\@title\par}
  \vspace{11pt}%
  \@author
  \vspace{8pt}%
  \papertitlerule\par
  \vspace{10pt}%
  \endgroup
  {\abscontent}%
  \thispagestyle{firststyle}}
\makeatother
\fancypagestyle{firststyle}{%
  \fancyhf{}%
  \renewcommand{\headrulewidth}{0pt}%
  \renewcommand{\footrulewidth}{0pt}%
  \fancyfoot[L]{\normalfont\fontsize{9}{11}\selectfont
    \textsuperscript{\papercontactmark}\,Corresponding author:
    \the\correspondingauthor}}

\begin{abstract}
Long-term problem-solving processes, such as scientific research and software development, proceed along an evolving trial-and-error trajectory toward a central goal.
We term these \textit{exploration tasks}, which require models to remember not only facts but also how solutions are tested against intermediate outcomes and early attempts are continually revised or revived.
However, existing long-term memory research focuses on factual recall and updates, overlooking the interconnected nature of exploration.
We observe that exploration processes naturally form graphs, with branching approaches linked by revisions and reconsiderations.
Motivated by this, we propose \textsc{LExMem}, a graph-guided benchmark designed to evaluate \underline{\textsc{Mem}}ory in \underline{\textsc{L}}ong-term \underline{\textsc{Ex}}ploration tasks.
We model exploration processes as graphs and introduce an evolutionary mechanism to increase their structural complexity, yielding 224 dialogues and 2,018 question-answer instances derived from graph structures.
Building on the same insight, we propose \textsc{LExGraph}, a graph-based memory framework that incrementally organizes dialogues into interconnected memory nodes and retrieves evidence through hybrid retrieval and neighbor expansion.
Extensive experiments demonstrate that \textsc{LExMem} poses a significant challenge to existing retrieval- and memory-based systems, whereas \textsc{LExGraph} achieves superior performance and generalizes effectively to external benchmarks.
Codes and datasets are available at \url{https://github.com/mbzuai-nlp/LExMem}.
\end{abstract}

\begin{document}
\raggedbottom
\maketitle

\section{Introduction}
Recent large language models (LLMs) have demonstrated impressive capabilities across a wide spectrum of tasks, including open-domain dialogue \citep{ICLR2024_b3075b88, xu-etal-2025-multiagentesc}, code generation \citep{10.1145/3744916.3773255}, and scientific research \citep{DBLP:journals/corr/abs-2506-01839, su-etal-2025-many}.
Long-term memory is essential for sustaining these interactions, enabling models to retain and utilize information accumulated over time.
Therefore, numerous benchmarks and frameworks have been developed to evaluate and enhance this capability~\citep{NEURIPS2024_6ddc001d, 3780338.3781176, 10.1609/aaai.v40i39.40644, xu-etal-2026-structmem}.
One line of research investigates memory for user personas and personal facts in extended conversations~\citep{DBLP:conf/acl/MaharanaLTBBF24, DBLP:conf/aaai/ZhongGGYW24, jiang2025know, ICLR2025_LongMemEval, ICLR2025_28a46044, li-etal-2026-locomo}.
In parallel, evaluations have expanded to task-oriented scenarios~\citep{ICLR2026_fd1eff9d, shen-etal-2026-mem2actbench}.
For instance, MemoryCode requires models to follow coding instructions that are introduced and updated across sessions~\citep{rakotonirina-etal-2025-tools}, while RealMem investigates whether agents can track the states of ongoing projects~\citep{bian-etal-2026-realmem}.
Collectively, existing benchmarks primarily assess the recall of information, ranging from personal facts to evolving task details.

\begin{figure}[t]
\centering
\includegraphics[width=0.6\textwidth]{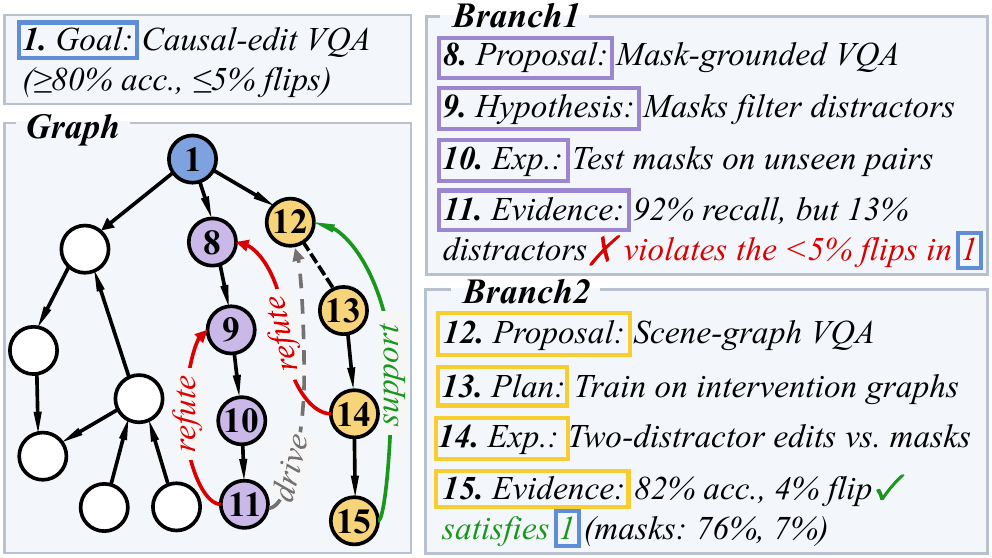}
\caption{
An illustrative exploration task modeled as a graph. We highlight two branches, whose exploration steps are interconnected by relations.
}
\label{fig:fig1}
\end{figure}

\begin{figure}[t]
\centering
\includegraphics[width=\textwidth]{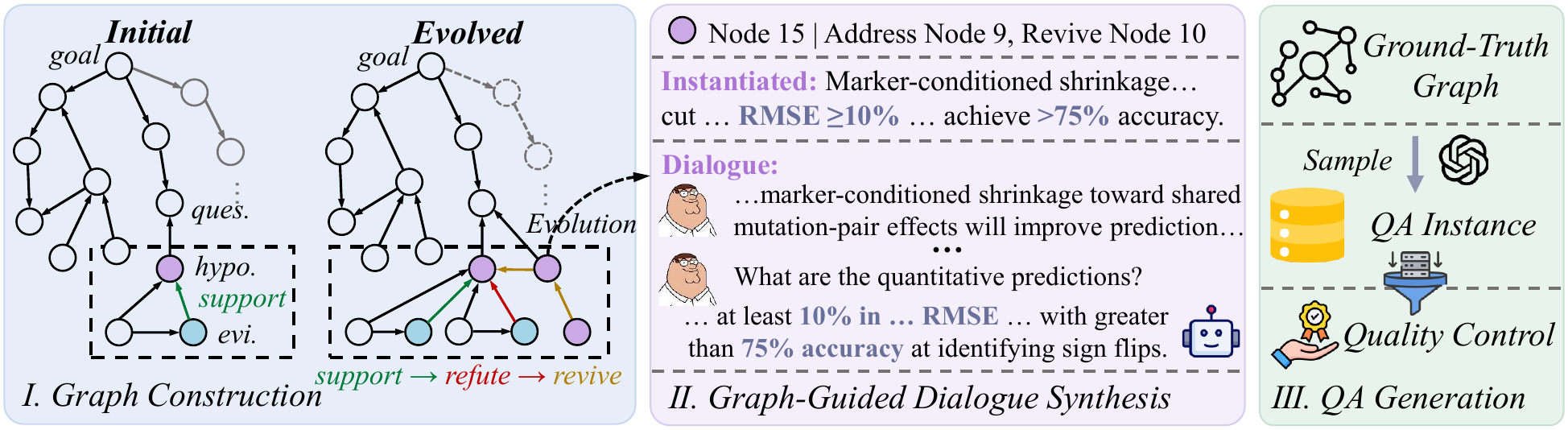}
\caption{
Overview of the \textsc{LExMem} construction pipeline. \textit{hypo.}, \textit{ques.}, and \textit{evi.} denote hypothesis, question, and evidence nodes, respectively, i.e., node types.
}
\label{fig:pipeline}
\end{figure}

However, real-world problem solving often involves \textit{exploration tasks}, where conversations evolve through iterative trial and error toward a central goal.
As illustrated in Figure~\ref{fig:fig1}, such tasks involve testing tentative approaches against empirical evidence and continually revising decisions based on intermediate outcomes.
Yet, existing benchmarks overlook the dynamics of exploration, leaving it unclear how effectively models can reason over interrelated information, trace the rationale of extended inquiry, and aggregate findings scattered across prolonged interactions~\citep{DBLP:conf/acl/MaharanaLTBBF24, ICLR2026_fd1eff9d, li-etal-2026-locomo}.
Furthermore, existing memory frameworks organize dialogue history around semantic similarity~\citep{NEURIPS2025_19909c36, ICLR2025_0382cb76}, topical clustering~\citep{tan-etal-2025-prospect, ICLR2026_a05b7265}, and entity relations~\citep{DBLP:conf/ecai/ChhikaraKASY25, edge2025localglobalgraphrag, guo-etal-2025-lightrag}, overlooking the procedural dependencies required in sustained exploration.

In this paper, we propose \textsc{LExMem}, a graph-guided benchmark for evaluating \underline{\textsc{Mem}}ory in \underline{\textsc{L}}ong-term \underline{\textsc{Ex}}ploration tasks.
To this end, we leverage directed graphs to model exploration steps and their dependencies, enabling the systematic construction of complex exploration trajectories and relationship-aware questions.
Specifically, we develop a three-stage construction pipeline.
First, we construct exploration graphs and introduce an evolutionary mechanism to increase their structural complexity.
Second, we instantiate these graphs with scientific research and software design tasks, and further expand each node into multi-turn dialogue blocks.
Finally, we extract structural subgraphs to generate QA instances that evaluate memory of interconnected exploration steps.
The resulting \textsc{LExMem} contains 224 dialogues and 2,018 QA instances across three graph-derived question types.
Extensive experiments reveal that \textsc{LExMem} poses substantial challenges to existing retrieval and memory frameworks, particularly in aggregating information scattered throughout exploration.

To address these challenges, we introduce \textsc{LExGraph}, a \underline{\textsc{Graph}}-based memory framework designed to maintain structured context in \underline{\textsc{L}}ong-term \underline{\textsc{Ex}}ploration.
Rather than relying on predefined graphs, \textsc{LExGraph} incrementally constructs a memory graph from raw dialogues, organizing information into nodes connected by hierarchical and cross-branch dependencies.
During inference, it retrieves relevant nodes and traverses their connections to integrate scattered evidence for reasoning.
Experiments demonstrate that \textsc{LExGraph} consistently outperforms existing retrieval and memory systems on \textsc{LExMem} and generalizes effectively to external memory benchmarks.

\section{Benchmark Curation}
\textsc{LExMem} is a graph-guided benchmark for evaluating long-term memory in exploration tasks with interconnected steps and dependencies, constructed via the pipeline illustrated in Figure \ref{fig:pipeline}.

\subsection{Evolutionary Graph Construction}
\label{sec:evolution_graph}
To capture the evolving dependencies in long-term exploration, we model each exploration process as a directed graph $G=(V,E)$, where nodes represent individual exploration steps and edges encode their relationships.
Specifically, $V$ consists of nine functional node types (e.g., artifact), while $E$ contains nine semantic relation types (e.g., support), as detailed in Appendix~\ref{sec:appendix_graph_evolution}.

\boldpara{Initial Graph Generation.}
We first prompt LLMs to generate an initial graph $G_0$ specifying only its topology (i.e., typed nodes and edges) without textual content, thereby separating structural generation from domain-specific content to explicitly control the complexity of exploration trajectories.
Each graph is rooted at a \textit{goal} node and hierarchically expands into downstream typed nodes, forming a spanning tree.
Additional edges connect nodes across branches to capture dependencies between exploration paths.
To model the dialogue progression, nodes are ordered chronologically via a depth-first preorder traversal, yielding a sequence of $N$ nodes $(u_1, \dots, u_N)$ with $u_1$ as the root.

\boldpara{Graph Evolution.}
While the initial graph establishes the dependencies among exploration steps, it may not adequately capture the non-monotonic nature of real-world exploration, where earlier approaches are revised, challenged, or revisited as new evidence emerges.
To simulate these dynamics, we introduce an evolutionary mechanism that iteratively modifies the initial graph $G_0$ according to predefined mutation rules (Appendix~\ref{sec:appendix_graph_evolution}).
Starting from $G=G_0$, we scan the node sequence $(u_1,\dots,u_N)$ in chronological order.
At each node $u_t$, we define the prefix subgraph $G_{\le t}$ as the visited nodes ${u_1,\dots,u_t}$ and their interconnecting edges.
If node $u_t$ and its incident edges match a rule, a mutation is triggered with probability $p$; when triggered, we prompt an LLM with $G_{\le t}$ to introduce new nodes and edges representing subsequent revisions, challenges, or reconsiderations.
This generation further extends to complete the rest of the graph, yielding an updated graph $G'$ in which $G_{\le t}$ remains unchanged.
We then set $G=G'$, update its node sequence accordingly, and resume scanning from $u_{t+1}$ until the entire graph is traversed.

Through LLM-based verification and manual curation, we retain 150 high-quality evolved graphs.
As shown in Table~\ref{tab:evolution_stats}, graph evolution substantially enriches the dynamics of sustained exploration while maintaining comparable graph sizes, providing a foundation for dialogue synthesis (\S\ref{sec:dialogue_synthesis}).

\begin{table}[t!]
\centering
\small
\setlength{\tabcolsep}{4pt}
\begin{adjustbox}{max width=0.95\columnwidth}
\begin{tabular}{lcc@{\hspace{2.8em}}}
\toprule
\textbf{Statistic (Avg. per graph)} & \textbf{Initial} & \textbf{Evolved} \\
\midrule
Nodes                                  & 58.3 & 61.0\inc[gray]{4.6} \\
Nodes w/ state changes                 & 1.97 & 4.19\inc{112.7} \\
Nodes w/ initial $\neq$ current state  & 1.82 & 3.81\inc{109.3} \\
\bottomrule
\end{tabular}
\end{adjustbox}
\caption{
Statistics of graphs before and after evolution.
For each node, we examine its incoming edges labeled with four relations (\textit{support}, \textit{refute}, \textit{revise}, \textit{revive}).
\textit{State change}: two consecutive edges have different relations; \textit{initial $\neq$ current}: the first and last edges differ.
\label{tab:evolution_stats}
}
\end{table}

\subsection{Graph-Guided Dialogue Synthesis}
\label{sec:dialogue_synthesis}

\boldpara{Seed Task Preparation.}
We first prepare 150 seed tasks across two domains: \textit{research exploration} (100) and \textit{software design} (50).
Each task specifies a primary objective along with several requirement metrics.
Details are provided in Appendix~\ref{sec:appendix_seed_task}.

\boldpara{Graph Instantiation.}
Since the evolved graphs (\S\ref{sec:evolution_graph}) specify only node types and edge relations without concrete text, we instantiate them with domain-specific tasks to create coherent exploration trajectories while preserving their underlying structures.
First, we randomly pair each seed task with four distinct evolved graphs, yielding 600 candidates, and directly assign a seed task with its task description and requirements to the root \textit{goal} node ($u_1$).
For each subsequent node $u_t$ ($t = 2, \dots, N$) in chronological order, we prompt the LLM with (1) all previously instantiated nodes and edges, (2) the type of $u_t$, and (3) the relations linking it to earlier nodes;
it then generates concrete descriptions for $u_t$ and its incident edges with specific details (e.g., exact metrics).
To ensure consistency, we conduct two stages of LLM-based validation: (1) step-wise checks against prior context (2.3\% filtered) and (2) a global consistency check over the completed graph (55.8\% filtered).
This finally yields 259 instantiated graphs, each denoted as $G^*$.

\boldpara{Dialogue Expansion.}
We further expand each instantiated graph $G^*$ into a natural multi-turn conversation, transforming structured exploration steps into information distributed across prolonged interactions.
Specifically, each node $u_t$ is expanded into a dialogue block of multiple user-assistant turns, with the LLM conditioned on (1) all preceding dialogue blocks, (2) the content of $u_t$, and (3) its relations to earlier nodes.
Each block must faithfully convey the corresponding node content without introducing unsupported facts, ensuring that the underlying exploration dependencies are preserved.
We apply an LLM-based faithfulness check after each block (0\% filtered), followed by a global dialogue consistency check (13.5\% filtered).
Finally, we obtain 224 high-quality dialogues with 51K to 174K tokens (88K on average), each denoted as $\mathcal{D}$ and paired with its instantiated graph $G^*$.

\subsection{QA Instance Generation}
\label{sec:qa_instance_generation}
To systematically evaluate memory of interconnected exploration processes, we derive three types of questions from distinct graph structures, each targeting a different reasoning capability.
Specifically, we extract candidate subgraphs or node sets $S \subseteq G^*$ following three patterns:
\begin{itemize}[leftmargin=*]
    \item \textbf{\textit{Dense subgraphs.}} To assess joint reasoning over tightly interrelated information, we sample connected subgraphs $S = (V_S, E_S)$ satisfying $|E_S| \ge 2|V_S|$ and $|V_S| \ge 4$, ranked by edge density with a Jaccard penalty to reduce overlap.
    \item \textbf{\textit{Sparse subgraphs.}} To evaluate reasoning along long-range exploration trajectories, we extract subgraphs where two distant nodes are linked solely by a multi-hop path passing through the root node, with no other edges among the path nodes. Candidates are similarly ranked by path length with a Jaccard penalty.
    \item \textbf{\textit{Same-type sets.}} To evaluate the aggregation of information scattered throughout the exploration process, we group all nodes sharing the same functional type into a set $S$.
\end{itemize}
By grounding each question in a specific graph structure, we can explicitly identify the supporting evidence and assess distinct forms of relational reasoning.
We sample up to nine candidates $S$ per pattern per graph, and prompt the LLM to formulate a question $q$ targeting $S$ alongside a concise answer set $\mathcal{A}$.
We then apply LLM-based verification to assess (1) question clarity and validity (39.3\% filtered) and (2) answerability based solely on $\mathcal{D}$ (23.9\% filtered).
This yields 2,018 high-quality QA instances, each formulated as $(\mathcal{D}, q, \mathcal{A})$, comprising 561 dense, 759 sparse, and 698 same-type instances.
We annotate their supporting evidence turns in Appendix~\ref{sec:appendix_evidence_annotation} and conduct a human evaluation of benchmark quality in Appendix~\ref{sec:appendix_human_evaluation}.

\section{Method}
\subsection{Overview}
We propose \textsc{LExGraph}, a graph-based memory framework that incrementally organizes evolving dialogue information and preserves dependencies among exploration steps.
Unlike the graph-guided benchmark construction, \textsc{LExGraph} builds its memory graph solely from raw dialogues, without access to the ground-truth graphs $G^*$ of \textsc{LExMem}.

First, given an ongoing dialogue $\mathcal{D}$, we incrementally construct a directed memory graph $\mathcal{G}=(\mathcal{M},\mathcal{L})$ with memory nodes $\mathcal{M}$ and relational edges $\mathcal{L}$ (\S\ref{sec:memory_graph_construction}).
Each node $m_i \in \mathcal{M}$ is a key-value pair $(k_i,v_i)$, where the key $k_i$ summarizes the content and key facts, while the value $v_i$ preserves the corresponding raw dialogue turns.
Second, given a question $q$, we retrieve relevant memory nodes and traverse their connections to aggregate supporting evidence for response generation (\S\ref{sec:memory_retrieval}).

\subsection{Memory Graph Construction}
\label{sec:memory_graph_construction}
\boldpara{Dialogue Segmentation.}
To organize conversational memory, we partition dialogue $\mathcal{D}$ into topic-coherent segments that serve as memory nodes.
Since users typically guide the discussion, we detect topic shifts based on user turns only.
Specifically, we maintain a buffer of unsegmented dialogue turns and invoke an LLM after every $n$ new user turns to identify topic boundaries within it.
Dialogue turns preceding the detected boundaries are grouped into memory nodes, each storing the corresponding user and assistant turns as its value $v_i$, while the remaining turns stay in the buffer.

\boldpara{Key Generation.}
For each node $m_i$, the LLM generates a structured key $k_i$ from its value $v_i$, comprising (1) a concise summary of the main discussion and (2) a list of salient facts recording concrete details, such as entities and numerical values.
This preserves both the overall context and concrete information that a general summary may omit.

\boldpara{Edge Construction.}
Once its key $k_i$ and value $v_i$ are obtained, we integrate node $m_i$ into the graph $\mathcal{G}$ by establishing two types of edges:
\begin{itemize}[leftmargin=*]
\item \textbf{Type}~\textbf{\uppercase\expandafter{\romannumeral1}} (Hierarchical): parent-child edges tracing the hierarchical progression of dialogue topics.
\item \textbf{Type}~\textbf{\uppercase\expandafter{\romannumeral2}} (Cross-branch): associative edges linking related information across different branches of the exploration process.
\end{itemize}

For \textbf{Type}~\textbf{\uppercase\expandafter{\romannumeral1}}, we draw on the \textit{right frontier constraint} from discourse theory \citep{asher2003logics} to guide parent selection.
It posits that in a discourse tree, a new utterance (here, node $m_i$) can only attach to the path from the root ($m_1$) to the most recent utterance ($m_{i-1}$), while utterances outside this path are closed to direct continuation.
We maintain this path as the right frontier $\mathcal{F}$ (initialized as $[m_1]$), which constrains parent selection and ensures structural coherence.
Given the keys of $m_i$ and $\mathcal{F}$, the LLM selects a parent $\mathrm{pa}(m_i) \in \mathcal{F}$ and describes their relation.
We then add the edge $(\mathrm{pa}(m_i), m_i)$ to $\mathcal{L}$, and update $\mathcal{F}$ to the new path $[m_1, \dots, \mathrm{pa}(m_i), m_i]$.

For \textbf{Type}~\textbf{\uppercase\expandafter{\romannumeral2}}, we consider historical nodes outside the updated frontier, defined as $\mathcal{H} = \{m_1, \dots, m_{i-1}\} \setminus \mathcal{F}$, to capture cross-branch associations, and select those relevant to $m_i$:
\begin{equation}
\mathcal{C}_i = \{ m_j \in \mathcal{H} \mid \cos(\mathbf{e}_j, \mathbf{e}_i) \ge \tau_{\mathrm{sim}} \},
\end{equation}
where $\mathbf{e}_i$ and $\mathbf{e}_j$ denote the key embeddings of $m_i$ and $m_j$, respectively, computed by \texttt{gte-Qwen2-1.5B} \citep{li2023generaltextembeddingsmultistage}, and $\tau_{\mathrm{sim}}$ is a threshold.
For each $m_j \in \mathcal{C}_i$, we prompt LLMs to determine their relation and edge direction, establishing an edge in $\mathcal{L}$.
Consequently, we transform the dialogue into an interconnected memory graph, which captures both the primary exploration trajectory and broader contextual dependencies.

\subsection{Memory Retrieval and Answering}
\label{sec:memory_retrieval}
\boldpara{Node Retrieval.}
Given a question $q$, we score each node $m_i \in \mathcal{M}$ through two complementary channels to capture both semantic relevance and lexical matches:
(1) \textbf{semantic matching}, measured by the cosine similarity $\cos(\mathbf{e}_q, \mathbf{e}_i)$ between $q$ and $k_i$, and (2) \textbf{lexical matching}, measured by the BM25 score between $q$ and $k_i$ \citep{10.1561/1500000019}.
To balance both retrieval signals, we fuse their individual rankings via Reciprocal Rank Fusion \citep{10.1145/1571941.1572114}:
\begin{equation}
s(m_i) = \frac{1}{\kappa + r_{\mathrm{sem}}(m_i)} + \frac{1}{\kappa + r_{\mathrm{lex}}(m_i)},
\end{equation}
where $r_{\mathrm{sem}}(m_i)$ and $r_{\mathrm{lex}}(m_i)$ are the semantic and lexical ranks of $m_i$, and $\kappa$ is a constant.
To reduce token consumption, we retrieve memory nodes via two steps:
(1) select the top-ranked nodes by $s(m_i)$ into a primary set $\mathcal{M}_{\mathrm{pri}}$ up to a budget $B_{\mathrm{pri}}$;
and (2) collect nodes directly connected to $\mathcal{M}_{\mathrm{pri}}$ into a neighbor set $\mathcal{M}_{\mathrm{nbr}}$, also ranked by $s(m_i)$, up to a budget $B_{\mathrm{nbr}}$.
This locates the primary evidence for question $q$ while gathering its connected context to capture relational associations.

\boldpara{Answer Generation.}
Finally, the LLM answers $q$ grounded in a context $X$ assembled from the retrieved memories, comprising:
(1) keys and values of primary nodes in $\mathcal{M}_{\mathrm{pri}}$, (2) keys of neighbor nodes in $\mathcal{M}_{\mathrm{nbr}}$, and (3) relation descriptions of edges connecting the retrieved nodes.
This grounds the answer in structured memories, thereby facilitating accurate and context-aware answering.

\begin{table}[t]
\centering
\small
\setlength{\tabcolsep}{3pt}
\providecommand{\ps}{\rlap{$^{*}$}}
\providecommand{\pss}{\rlap{$^{**}$}}
\newcolumntype{N}{w{c}{3.4em}}
\newcolumntype{L}{>{\hspace*{5pt}\hbox to 3.4em\bgroup\hss}l<{\hss\egroup}}
\begin{adjustbox}{max width=\textwidth}
\begin{tabular}{c@{\hspace{10pt}}l*{5}{LN}}
\toprule
\multirow{2}{*}[-0.7ex]{\textbf{Model}} & \multirow{2}{*}[-0.7ex]{\textbf{Method}}
& \multicolumn{2}{>{\hspace*{5pt}}c}{\textbf{Dense}}
& \multicolumn{2}{>{\hspace*{5pt}}c}{\textbf{Sparse}}
& \multicolumn{2}{>{\hspace*{5pt}}c}{\textbf{Same-Type}}
& \multicolumn{2}{>{\hspace*{5pt}}c}{\textbf{Overall}}
& \multicolumn{2}{>{\hspace*{5pt}}c}{\textbf{Efficiency}} \\
\cmidrule(l{8pt}r){3-4} \cmidrule(l{8pt}r){5-6} \cmidrule(l{8pt}r){7-8} \cmidrule(l{8pt}r){9-10} \cmidrule(l{8pt}r){11-12}
& & \textbf{F1} & \textbf{ACC} & \textbf{F1} & \textbf{ACC} & \textbf{F1} & \textbf{ACC} & \textbf{F1} & \textbf{ACC}
& \textbf{Tok.$\downarrow$} & \textbf{Lat.$\downarrow$} \\
\midrule
& Vanilla    & 0.565 & 65.74 & 0.570 & 61.64 & 0.599 & 48.86 & 0.579 & 58.36 & \rx{1.0} & \rx{1.0} \\
& HippoRAG   & 0.587 & 67.26 & 0.576 & 62.12 & 0.614 & 47.69 & 0.592 & 58.56 & \rx{1.3} & \rx{1.1} \\
& ComoRAG    & 0.611 & 68.59 & 0.611 & 65.99 & 0.625 & 50.00 & 0.616 & 61.18 & \rx{1.8} & \rx{3.4} \\
& HippoRAG2  & 0.580 & 68.36 & 0.588 & 64.82 & 0.616 & 50.43 & 0.595 & 60.83 & \rx{1.2} & \rx{1.1} \\
& StructMem  & 0.621 & 66.30 & 0.641 & 66.22 & 0.638 & 48.08 & 0.634 & 59.97 & \rx{1.4} & \rx{1.1} \\
& GraphRAG   & 0.616 & 70.59 & 0.616 & \textbf{67.22} & 0.629 & 49.23 & 0.620 & 61.93 & \rx{1.6} & \rx{1.1} \\
& Mem0       & 0.614 & 68.59 & 0.595 & 67.05 & 0.635 & 49.62 & 0.614 & 61.45 & \rx{6.7} & \rx{1.1} \\
& LightMem   & 0.602 & 68.30 & 0.606 & 65.28 & 0.648 & 51.92 & 0.619 & 61.50 & \rx{1.1} & \rx{1.0} \\
& LightRAG   & 0.619 & 67.97 & 0.611 & 67.05 & 0.643 & 50.77 & 0.624 & 61.67 & \rx{1.6} & \rx{1.2} \\
\rowcolor{oursbg}
\cellcolor{white}\multirow{-10}{*}{\rotatebox[origin=c]{90}{Qwen-3.8 27B}}
& \textsc{LExGraph} & \textbf{0.658}\pss & \textbf{70.89}\ps & \textbf{0.665}\pss & 66.30 & \textbf{0.676}\ps & \textbf{55.73}\pss & \textbf{0.667}\pss & \textbf{63.92}\ps & \rx{1.3} & \rx{1.1} \\
\midrule
& Vanilla    & 0.524 & 63.72 & 0.459 & 60.37 & 0.526 & 53.81 & 0.500 & 59.03 & \rx{1.0} & \rx{1.0} \\
& HippoRAG   & 0.528 & 63.42 & 0.476 & 62.90 & 0.506 & 55.35 & 0.501 & 60.43 & \rx{1.3} & \rx{1.1} \\
& ComoRAG    & 0.529 & 67.17 & 0.468 & 59.15 & 0.513 & 55.40 & 0.501 & 60.08 & \rx{1.5} & \rx{3.8} \\
& HippoRAG2  & 0.548 & 69.37 & 0.496 & 64.50 & 0.527 & 56.12 & 0.521 & 62.97 & \rx{1.2} & \rx{1.0} \\
& StructMem  & 0.539 & 64.72 & 0.478 & 63.25 & 0.546 & 57.91 & 0.519 & 61.81 & \rx{1.3} & \rx{0.8} \\
& GraphRAG   & 0.555 & 69.82 & 0.473 & 63.49 & 0.533 & 56.17 & 0.516 & 62.72 & \rx{1.7} & \rx{0.9} \\
& Mem0       & 0.567 & 68.79 & 0.463 & 64.71 & 0.567 & 56.79 & 0.528 & 63.10 & \rx{9.1} & \rx{1.2} \\
& LightMem   & 0.559 & 68.52 & 0.487 & \textbf{66.61} & 0.529 & 56.34 & 0.522 & 63.59 & \rx{1.2} & \rx{0.8} \\
& LightRAG   & 0.551 & 69.59 & 0.492 & 64.44 & 0.546 & 57.44 & 0.527 & 63.46 & \rx{1.6} & \rx{1.2} \\
\rowcolor{oursbg}
\cellcolor{white}\multirow{-10}{*}{\rotatebox[origin=c]{90}{GPT-6 Luna}}
& \textsc{LExGraph} & \textbf{0.569}\ps & \textbf{70.02} & \textbf{0.535}\pss & 64.97 & \textbf{0.598}\pss & \textbf{58.12} & \textbf{0.566}\pss & \textbf{64.00}\ps & \rx{1.2} & \rx{1.1} \\
\bottomrule
\end{tabular}
\end{adjustbox}
\caption{
Results on \textsc{LExMem}.
Tok. and Lat. represent the average token usage and latency (retrieval plus answering) per question, relative to Vanilla (1.0$\times$).
For \texttt{Qwen-3.8 27B}, 1.0$\times$ corresponds to 102.9K tokens and 38.5 seconds; 
for \texttt{GPT-6 Luna}, 95.3K tokens and 8.22 seconds. 
Best results are in bold.
$^{*}p<0.05$ and $^{**}p<0.01$ indicate statistically significant over the LightRAG baseline, the strongest baseline on average, using paired bootstrap.
}
\label{tab:main_results}
\end{table}

\section{Experimental Setting}
\subsection{Baselines}
Each instance in \textsc{LExMem} is formulated as $(\mathcal{D}, q, \mathcal{A})$ (\S\ref{sec:qa_instance_generation}).
Given dialogue $\mathcal{D}$ and question $q$, each baseline is required to generate an answer $\hat{\mathcal{A}}$.
To ensure a comprehensive analysis, we compare \textsc{LExGraph} with the following baselines:
(1) \textbf{Retrieval-based methods,} which build indices or structured representations over the context to retrieve information, including ComoRAG \citep{10.1609/aaai.v40i39.40644}, GraphRAG \citep{edge2025localglobalgraphrag}, HippoRAG \citep{NEURIPS2024_6ddc001d}, HippoRAG2 \citep{3780338.3781176}, and LightRAG \citep{guo-etal-2025-lightrag};
and (2) \textbf{Memory-based methods,} which construct and manage memory structures, including LightMem \citep{ICLR2026_a05b7265}, Mem0 \citep{DBLP:conf/ecai/ChhikaraKASY25}, and StructMem \citep{xu-etal-2026-structmem}.
For these approaches, we first build their own index or memory over $\mathcal{D}$, and then retrieve context relevant to $q$, on which the backbone LLM conditions to generate $\hat{\mathcal{A}}$.
Details are described in Appendix~\ref{sec:appendix_details_of_baselines}.

\subsection{Benchmarks and Evaluation Metrics}
\label{sec:evaluation_metrics}
We conduct comprehensive evaluations on \textsc{LExMem} with two metrics:
(1) \textbf{Token-level F1}, adopted from SQuAD~\citep{rajpurkar-etal-2016-squad} to measure lexical overlap between $\hat{\mathcal{A}}$ and $\mathcal{A}$;
and (2) \textbf{Accuracy (ACC)}, following BEAM~\citep{ICLR2026_d7f0cfa0}, where an LLM judge scores each answer as correct ($1.0$), partially correct ($0.5$), or incorrect ($0.0$).
To verify the generalizability of our method \textsc{LExGraph}, we further evaluate it on the 128K-token subset of BEAM~\citep{ICLR2026_d7f0cfa0}, which covers ten long-term conversational memory abilities, and report ACC.

\subsection{Implementation Details}
We use \texttt{GPT-6 Sol}~\citep{openai_gpt6sol} for benchmark curation.
For evaluation, we employ \texttt{Qwen-3.8 27B} and \texttt{GPT-6 Luna}~\citep{openai_gpt6luna} as backbone models, with \texttt{gte-Qwen2-1.5B} \citep{li2023generaltextembeddingsmultistage} as the embedding model and \texttt{GPT-4o mini}~\citep{openai_gpt4omini} as the ACC evaluator.
We conduct experiments on two RTX 5880 GPUs with seed 42 and temperature 0.
Details are in Appendix~\ref{sec:appendix_implementation_details}.

\section{Experiments}

\subsection{Main Results}
We present the evaluation results on \textsc{LExMem} in Table \ref{tab:main_results}.
These results reveal several key observations.
First, retrieval- and memory-based frameworks consistently outperform the Vanilla baseline in overall metrics.
This demonstrates that augmenting LLMs with external retrieval or memory is essential to capture the evolving, interconnected contexts and dependencies in exploration tasks.
Second, \textsc{LExGraph} achieves the highest overall performance across different backbones, outperforming strong baselines such as GraphRAG and LightMem.
This illustrates the effectiveness of our method, which can be attributed to structuring memory and capturing relational connections along the exploration trajectories.
Third, while most methods yield better results on \textit{Dense} and \textit{Sparse} sets, their accuracy drops substantially on the \textit{Same-Type} set across backbones.
For example, on \texttt{GPT-6 Luna}, LightMem achieves an accuracy of 68.52 on \textit{Dense} and 66.61 on \textit{Sparse} sets, yet its accuracy falls to 56.34 on \textit{Same-Type} set.
This may stem from the intrinsic difficulty of its question setting, which requires models to identify and aggregate multiple exploration steps scattered across the dialogue.
In contrast, our method maintains a solid advantage across all subsets, demonstrating its effectiveness in long-term problem-solving tasks.
Fourth, our method also achieves a balance between performance and efficiency, sustaining high accuracy with competitive resource overhead.

\begin{figure}[t]
\centering
\includegraphics[width=\textwidth]{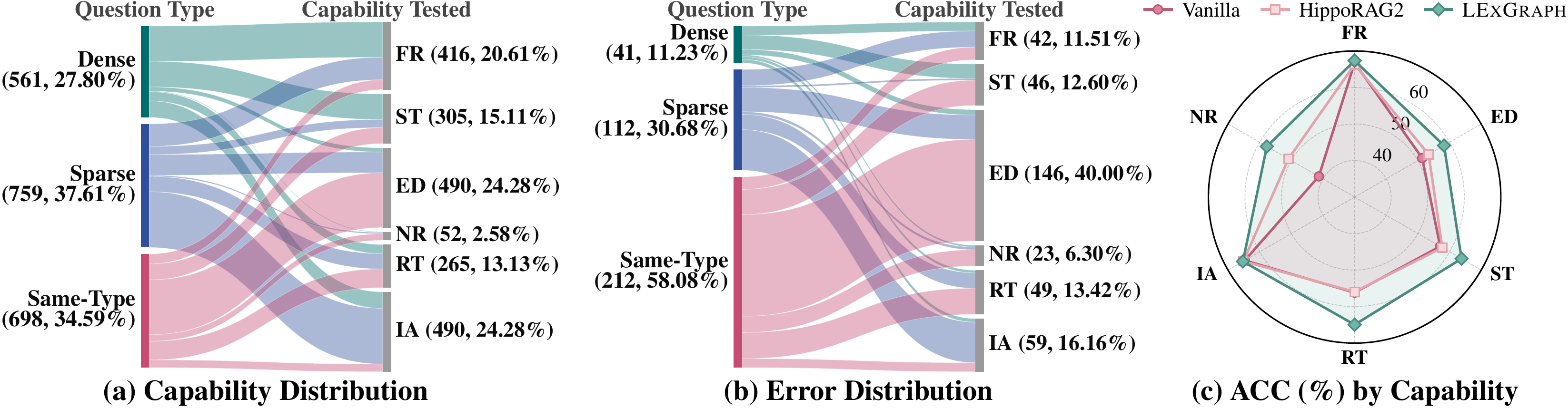}
\caption{
Capability distribution of \textsc{LExMem} by question type (a), together with Vanilla's errors (b) and per-capability accuracy of three methods (c) using \texttt{Qwen-3.8 27B}.
}
\label{fig:acc_sankey_radar}
\end{figure}

\subsection{Benchmark Analysis}
\boldpara{Capability Distribution.}
To enable fine-grained evaluation on exploration tasks, we categorize all instances into six capabilities: Fact Retrieval (FR), State Tracking (ST), Entity Discrimination (ED), Numeric Reasoning (NR), Relation Tracing (RT), and Information Aggregation (IA), with details of definitions and categorization in Appendix~\ref{sec:appendix_benchmark_capability}.
As shown in Figure~\ref{fig:acc_sankey_radar}(a), each question type exhibits distinct capability focuses:
\textit{Dense} questions primarily map to FR and ST, indicating that tightly coupled contexts mainly test localized facts and state changes;
\textit{Sparse} questions are dominated by IA, which demands evidence aggregation across long-range exploration trajectories;
and \textit{Same-Type} questions overwhelmingly target ED, requiring fine-grained discrimination among similar entities.

\boldpara{Error Analysis.}
To diagnose failures, we analyze instances where the Vanilla baseline achieves zero accuracy on \textsc{LExMem}, as shown in Figure~\ref{fig:acc_sankey_radar}(b).
First, errors are mainly concentrated in \textit{Same-Type} questions and dominated by ED, indicating that models may easily confuse similar contexts accumulated during exploration.
\textit{Same-Type} questions also account for nearly all failures in NR, suggesting that the primary difficulty of numerical reasoning lies in comparative reasoning across similar information.
Second, \textit{Dense} errors mainly consist of FR and ST, consistent with the primary capabilities tested by \textit{Dense} questions.
Third, \textit{Sparse} errors primarily flow into IA while spanning capabilities such as FR, ED, and RT.
This demonstrates that long-term exploration presents compound reasoning challenges where extended and evolving contexts disrupt global evidence synthesis.

\boldpara{Performance by Capability.}
We compare the accuracy across six capabilities in Figure~\ref{fig:acc_sankey_radar}(c).
While HippoRAG2 achieves a notable improvement on NR, its gains on other dimensions remain marginal.
In contrast, \textsc{LExGraph} consistently outperforms both baselines, with the most pronounced advantages on NR and RT.
This may stem from \textsc{LExGraph}'s hierarchical and cross-branch edges that ensure multi-hop tracing, alongside structured keys enabling comparative reasoning.

\begin{figure}[t]
\centering
\captionsetup{justification=centering,singlelinecheck=true}
\begin{minipage}[c]{0.38\textwidth}
\centering
\includegraphics[width=\linewidth]{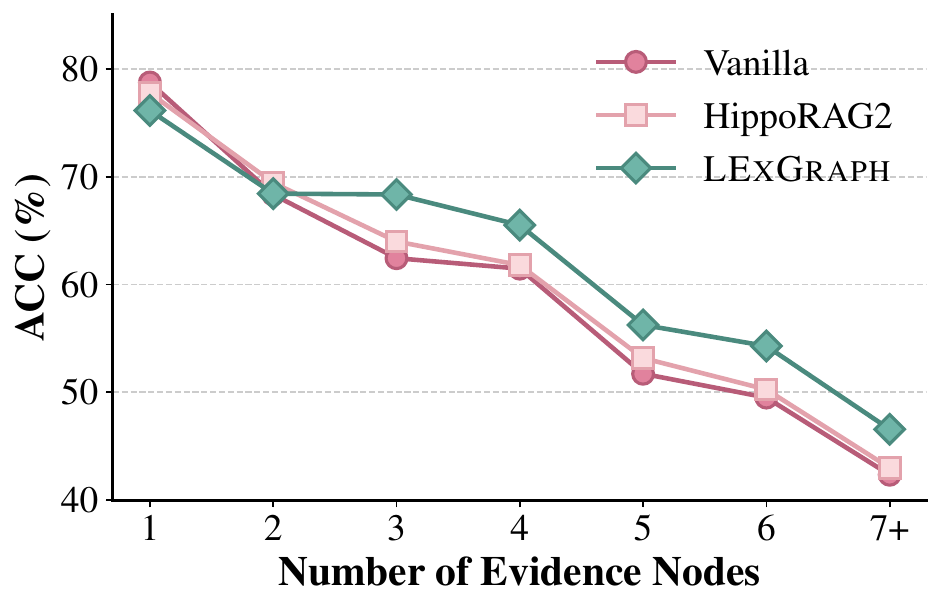}
\end{minipage}\hspace{0.05\textwidth}
\begin{minipage}[c]{0.50\textwidth}
\centering
\footnotesize
\setlength{\tabcolsep}{4pt}
\begin{adjustbox}{max width=\linewidth}
\begin{tabular}{lcccc}
\toprule
\textbf{Memory Ability} & Vanilla & Mem0 & LightRAG & \textsc{LExGraph} \\
\midrule
Abstention               & 0.500 & 0.525 & 0.488 & \textbf{0.650} \\
Contradiction Resolution & 0.181 & 0.169 & 0.200 & \textbf{0.216} \\
Event Ordering           & 0.167 & 0.176 & 0.194 & \textbf{0.218} \\
Information Extraction   & 0.644 & 0.633 & \textbf{0.695} & 0.656 \\
Instruction Following    & 0.394 & 0.481 & 0.500 & \textbf{0.562} \\
Knowledge Update         & 0.375 & 0.493 & 0.481 & \textbf{0.525} \\
Multi-Hop Reasoning      & 0.422 & \textbf{0.493} & 0.441 & 0.451 \\
Preference Following     & 0.613 & 0.648 & \textbf{0.725} & 0.721 \\
Summarization            & 0.359 & 0.385 & 0.320 & \textbf{0.389} \\
Temporal Reasoning       & 0.300 & 0.325 & 0.313 & \textbf{0.331} \\
\midrule
Overall                  & 0.396 & 0.433 & 0.436 & \textbf{0.472} \\
\bottomrule
\end{tabular}
\end{adjustbox}
\end{minipage}

\begin{minipage}[t]{0.38\textwidth}
\captionof{figure}{
Accuracy on \textsc{LExMem} across different numbers of evidence nodes, using \texttt{Qwen-3.8 27B}.
}
\label{fig:acc_by_evidence_nodes}
\end{minipage}\hspace{0.05\textwidth}
\begin{minipage}[t]{0.50\textwidth}
\captionof{table}{Results on BEAM. Best results are in bold.}
\label{tab:beam_results}
\end{minipage}
\end{figure}

\begin{figure}[t]
\centering
\includegraphics[width=\textwidth]{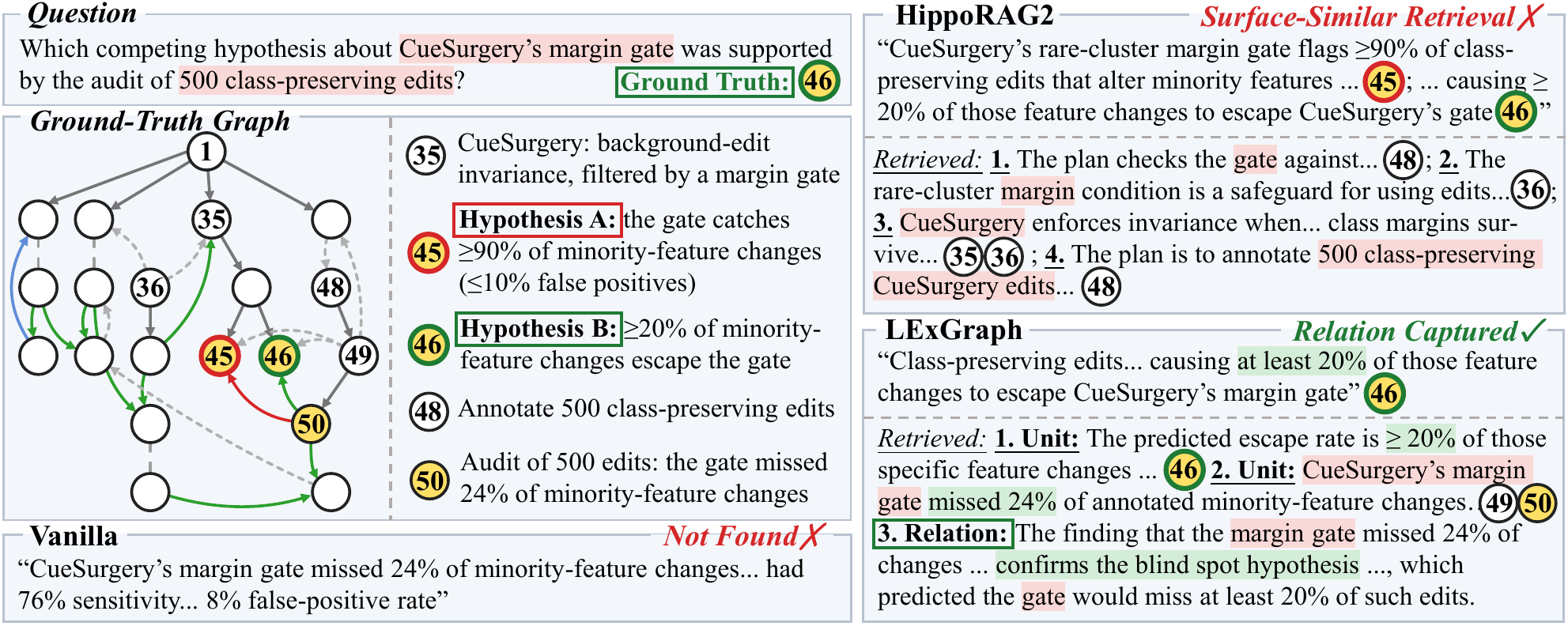}
\caption{
A qualitative example from \textsc{LExMem} with \texttt{Qwen-3.8 27B}.
Vanilla fails to relate the key finding to the hypotheses, and HippoRAG2 retrieves only surface-similar content, whereas \textsc{LExGraph} captures their relation.
\hlc{evidyellow}{Yellow}: evidence nodes; \hlc{simred}{red}: content similar to the question; \hlc{oursgreen}{green}: key evidence and relations retrieved by \textsc{LExGraph}.
In the ground-truth graph $G^*$, green/red edges denote \textit{support}/\textit{refute}, and undirected dashed edges denote omitted intermediate steps.
Each text is tagged with the node it comes from.
}
\label{fig:case_study}
\end{figure}

\subsection{Method Analysis}
\boldpara{Effect of Evidence Count.}
While each question is generated from a subgraph or node set $S \subseteq G^*$, only a subset of nodes in $G^*$ actually contains the supporting evidence (annotated in Appendix~\ref{sec:appendix_evidence_annotation}).
We group instances by the number of these ground-truth evidence nodes.
As shown in Figure~\ref{fig:acc_by_evidence_nodes}, accuracy across all methods declines steadily as the number of evidence nodes increases, suggesting the challenge of aggregating evidence scattered across long-term exploration tasks.
While Vanilla and HippoRAG2 perform competitively when evidence is concentrated in fewer nodes, their accuracy drops sharply as the evidence becomes increasingly scattered across the dialogue.
In contrast, \textsc{LExGraph} maintains a clear advantage over both baselines when synthesizing such dispersed evidence.

\boldpara{Results on the External Benchmark.}
We further assess \textsc{LExGraph} on BEAM~\citep{ICLR2026_d7f0cfa0}, which evaluates long-term conversational memory across multiple capabilities.
As shown in Table~\ref{tab:beam_results}, \textsc{LExGraph} achieves the top performance in seven out of ten memory abilities, yielding the highest overall score of 0.472 over strong baselines such as LightRAG.
In particular, it demonstrates pronounced advantages in \textit{Abstention} and \textit{Instruction Following}, alongside consistent strengths in state-tracking tasks such as \textit{Knowledge Update} and \textit{Event Ordering}.
These results indicate that our method generalizes effectively to broader conversational memory scenarios beyond exploration tasks.
We also conduct an oracle analysis in Appendix~\ref{sec:appendix_oracle} and an ablation study in Appendix~\ref{sec:appendix_ablation}.

\subsection{Case Study}
We present a case study in Figure~\ref{fig:case_study}, illustrating how each method responds when reasoning over interconnected exploration steps.
In this case, an intermediate step introduces CueSurgery's margin gate and branches into two competing hypotheses (nodes 45, 46).
A subsequent audit on a separate branch reveals that the gate missed 24\% of targeted changes (nodes 48, 50), the crucial evidence that must be associated with the earlier hypotheses in memory to resolve the question.
Vanilla recalls the isolated 24\% miss rate but fails to relate it to either hypothesis.
As HippoRAG2 relies on semantic similarity, it overlooks interrelated exploration steps and retrieves keyword-matched context instead of the essential evidence (node 50), leading to failure.
In contrast, \textsc{LExGraph} accurately locates the relevant evidence and captures the explicit relation across the exploration, thereby reaching the correct conclusion.
This demonstrates the effectiveness of \textsc{LExGraph} in preserving dependencies among exploration steps for context-aware reasoning.

\section{Related Work}
\subsection{Long-Term Memory Benchmarks}
In long-term memory evaluation, early research focuses on open-domain conversations, evaluating models' ability to retain and recall personal facts across sessions~\citep{DBLP:conf/acl/MaharanaLTBBF24, jiang2025know}.
Subsequent work broadens this scope to assess more fine-grained capabilities, including temporal reasoning and knowledge updates~\citep{ICLR2025_LongMemEval}, preference adherence~\citep{ICLR2025_28a46044}, and cognitive memory~\citep{li-etal-2026-locomo}.
Other efforts further incorporate multimodal contexts~\citep{bei-etal-2026-mem} and evaluate hallucinations within memory systems~\citep{chen2026halumemevaluatinghallucinationsmemory}.
A parallel line of research concentrates on long-context benchmarks, assessing comprehension and retrieval abilities over inputs scaling from thousands to millions of tokens~\citep{bai-etal-2024-longbench, bai-etal-2025-longbench, ICLR2026_d7f0cfa0}.
More recently, memory evaluation has expanded toward task-oriented dialogues, centering on task-relevant information such as user instructions and task progress, which accumulates and updates as the conversation unfolds~\citep{rakotonirina-etal-2025-tools, bian-etal-2026-realmem, shen-etal-2026-mem2actbench}.
However, existing benchmarks focus primarily on recalling isolated facts or tracking monotonic updates, overlooking the evolving process of exploration and its scattered, interconnected information.

\subsection{Memory and Retrieval Frameworks}
Early memory frameworks manage dialogue history as independent memory units, such as summaries and memos~\citep{lu2023memochattuningllmsuse, DBLP:conf/aaai/ZhongGGYW24, DBLP:conf/ecai/ChhikaraKASY25}.
Further efforts introduce management strategies such as reflective mechanisms~\citep{tan-etal-2025-prospect}, hierarchical storage~\citep{kang-etal-2025-memory}, and memory compression~\citep{ICLR2026_a05b7265} to balance accuracy and efficiency.
To capture associations among memory units, subsequent studies organize conversational history into structured representations, including trees~\citep{ICLR2025_0382cb76}, graphs~\citep{NEURIPS2025_19909c36, rasmussen2025zeptemporalknowledgegraph, xu-etal-2026-structmem}, and hypergraphs~\citep{yue-etal-2026-hypermem}.
Another line of research on retrieval-augmented generation (RAG) structures the corpus into hierarchical summary trees~\citep{ICLR2024_8a2acd17, 10.1609/aaai.v40i39.40644} or entity knowledge graphs~\citep{NEURIPS2024_6ddc001d, edge2025localglobalgraphrag, guo-etal-2025-lightrag, 3780338.3781176} to facilitate multi-hop retrieval.
However, current frameworks primarily capture semantic or entity-level associations rather than dialogue progression, where exploration branches out, revisits earlier attempts, and builds upon intermediate outcomes.

\section{Conclusion}
In this paper, we address the shortcomings of evaluating and enhancing long-term memory in exploration tasks.
We propose \textsc{LExMem}, a graph-guided benchmark that models exploration processes as graphs enriched through evolution, deriving 224 dialogues and 2,018 QA instances from graph structures.
We further present \textsc{LExGraph}, a graph-based memory framework that structures dialogues into memory graphs and gathers evidence by hybrid retrieval and neighbor expansion.
Extensive experiments demonstrate that existing retrieval and memory frameworks struggle on \textsc{LExMem}, whereas \textsc{LExGraph} achieves consistent improvements over strong baselines and generalizes to external benchmarks.
By providing a systematic evaluation benchmark and an effective memory framework, this work helps develop more reliable conversational agents for long-term exploration.

\section*{Limitations}
Despite the contributions of \textsc{LExMem} and \textsc{LExGraph}, our study is subject to several limitations that remain to be addressed in future research.
\boldpara{Synthetic Dialogues.}
Real-world exploration dialogues are difficult to collect at scale, as they span long horizons, often involve unpublished research or proprietary development, and lack annotations of the dependencies among exploration steps.
We therefore synthesize dialogues from instantiated exploration graphs, which allows explicit control over trajectory complexity and provides ground-truth structures for question generation, with quality verified through LLM-based checks and human evaluation.
Nevertheless, synthetic dialogues may not fully capture the stylistic diversity of real conversations.

\boldpara{Evaluation of Downstream Exploration.}
LExMem evaluates memory of the exploration process via question answering, rather than whether stronger memory leads to better exploration decisions. 
Such end-to-end evaluation is inherently difficult, as open-ended exploration rarely has a unique correct next step, and verifying one would require actually conducting the corresponding experiments. We leave this to future work.

\bibliography{reference}

\clearpage
\appendix
\setcounter{topnumber}{2}
\setcounter{totalnumber}{3}

\section{Details of Benchmark Curation}
\label{sec:appendix_benchmark_construction}

\subsection{Details of Evolutionary Graph Construction}
\label{sec:appendix_graph_evolution}
In this paper, $V$ contains nodes of nine functional types: \textit{goal}, \textit{proposal}, \textit{question}, \textit{hypothesis}, \textit{plan}, \textit{activity}, \textit{evidence}, \textit{decision}, and \textit{artifact}; $E$ consists of edges labeled with nine relations: \textit{address}, \textit{propose}, \textit{depend-on}, \textit{decompose}, \textit{evaluate}, \textit{support}, \textit{refute}, \textit{revise}, and \textit{revive}.

As outlined in \S\ref{sec:evolution_graph}, we define six mutation rules grouped into five categories to introduce realistic exploration twists.
Each rule specifies a structural precondition in the prefix subgraph $G_{\le t}$ and a required follow-up evolutionary event.
We detail the categories below.

\boldpara{Contradictory Evidence.}
Real-world exploration frequently uncovers conflicting signals rather than smooth consensus.
This rule matches when an \textit{evidence} node supports a \textit{hypothesis} or \textit{proposal}, and no \textit{refute} edge points to that target in $G_{\le t}$.
We specify that subsequent nodes should introduce new \textit{evidence} that refutes the target while preserving the existing \textit{support} edge, thereby capturing an emergent discussion.

\boldpara{Reconsidering Refuted Ideas.}
As investigation deepens, prematurely dismissed concepts are sometimes vindicated.
This rule applies when a \textit{hypothesis} or \textit{proposal} is refuted by \textit{evidence} and has received no incoming \textit{support} edge in $G_{\le t}$.
In this case, subsequent nodes should introduce new \textit{evidence} that supports the target, while keeping the earlier refutation.

\boldpara{Revising Hypotheses.}
Hypothesis adjustments often prompt chained modifications or a return to earlier assumptions.
This category matches when a \textit{hypothesis} node revises an earlier one, branching into two rules:
(1) a later \textit{hypothesis} modifies the new hypothesis once again;
(2) new evidence refutes the new \textit{hypothesis} and supports the earlier one, provided the new \textit{hypothesis} has not yet been refuted.
This second trajectory captures realizing an earlier idea was correct after a failed modification, while preserving the intermediate attempt in the graph history.

\boldpara{Revising Artifacts.}
System components such as datasets typically undergo iterative refinement before deployment.
This rule matches when an \textit{artifact} revises an earlier \textit{artifact}.
The subsequent graph is encouraged to introduce an additional \textit{artifact} that revises it once again, ensuring that intermediate artifacts represent ongoing development rather than final deployment.

\boldpara{Evaluating Revived Ideas.}
This category matches when an \textit{evidence} or \textit{decision} node revives an earlier \textit{hypothesis} or \textit{proposal} via a \textit{revive} edge, and the target currently lacks follow-up evaluation in $G_{\le t}$.
The follow-up trajectory must resolve the revived thread by introducing new \textit{evidence} that either validates or refutes it.

\subsection{Details of Seed Task Preparation}
\label{sec:appendix_seed_task}
We construct seed tasks on solution exploration across two domains: \textit{research exploration} and \textit{software design}.
Given a domain, we prompt LLMs to generate a concise task description that defines the primary objective, paired with one to three requirements that specify concrete metrics.
We then filter out low-quality tasks using LLMs, followed by manual review, yielding 150 seed tasks (100 for research exploration and 50 for software design).

\subsection{Evidence Annotation}
\label{sec:appendix_evidence_annotation}
To support the evaluation of evidence retrieval (Appendix~\ref{sec:appendix_retrieval}), we annotate supporting evidence for each QA instance.
Specifically, when generating question $q$ and answer set $\mathcal{A}$, the LLM also annotates the minimal set of evidence turns $\mathcal{E} \subseteq \mathcal{D}$ required to answer $q$.
During verification, in addition to the two checks in \S\ref{sec:qa_instance_generation}, the LLM examines whether $\mathcal{E}$ is both sufficient and necessary for answering $q$, filtering out 4.4\% of the instances.
Each resulting instance is thus formulated as $(\mathcal{D}, q, \mathcal{A}, \mathcal{E})$, with an average of 4.12 evidence turns per instance.

\subsection{Human Evaluation}
\label{sec:appendix_human_evaluation}
To verify the quality of \textsc{LExMem}, we conduct a human evaluation of three levels: instantiated graphs, dialogue blocks, and QA instances.
For each level, we randomly sample 75 items, where each dialogue block corresponds to the multi-turn dialogue expanded from a single node.
We employ two graduate students with research backgrounds to independently rate each item on a 1--5 scale along three criteria: (1) \textit{Coherence}, \textit{Exploration Fidelity}, and \textit{Domain Plausibility} for graphs; (2) \textit{Faithfulness}, \textit{Naturalness}, and \textit{Coherence} for dialogue blocks; and (3) \textit{Clarity}, \textit{Answerability}, and \textit{Evidence Validity} for QA instances.
As shown in Table~\ref{tab:human_evaluation}, all criteria receive high average scores with reliable inter-annotator agreement, which demonstrates the high quality of our benchmark.

\begin{table}[t]
\centering
\small
\begin{tabular}{llcc}
\toprule
\textbf{Level} & \textbf{Criterion} & \textbf{Avg.} & {\boldmath\textbf{$\rho$}} \\
\midrule
\multirow{3}{*}{Graph} & Coherence & 4.10 & 0.58 \\
 & Exploration Fidelity & 4.21 & 0.52 \\
 & Domain Plausibility & 4.42 & 0.61 \\
\midrule
\multirow{3}{*}{Block} & Faithfulness & 4.60 & 0.53 \\
 & Naturalness & 4.46 & 0.51 \\
 & Coherence & 4.27 & 0.58 \\
\midrule
\multirow{3}{*}{QA} & Clarity & 4.37 & 0.66 \\
 & Answerability & 4.83 & 0.59 \\
 & Evidence Validity & 4.69 & 0.67 \\
\bottomrule
\end{tabular}
\caption{Human evaluation results. Avg. denotes the average score of the two annotators; $\rho$ is Spearman's correlation between them.}
\label{tab:human_evaluation}
\end{table}

\begin{table}[t]
\centering
\small
\begin{adjustbox}{max width=\columnwidth}
\begin{tabular}{lcccccc}
\toprule
\multirow{2}{*}[-0.5ex]{\textbf{Method}}
& \multicolumn{2}{c}{\textbf{Metrics@5}}
& \multicolumn{2}{c}{\textbf{Metrics@10}}
& \multicolumn{2}{c}{\textbf{Metrics@20}} \\
\cmidrule(lr){2-3} \cmidrule(lr){4-5} \cmidrule(lr){6-7}
& \textbf{Recall} & \textbf{NDCG} & \textbf{Recall} & \textbf{NDCG} & \textbf{Recall} & \textbf{NDCG} \\
\midrule
GraphRAG   & 0.137 & 0.120 & 0.206 & 0.150 & 0.251 & 0.166 \\
HippoRAG   & 0.175 & 0.136 & 0.282 & 0.179 & 0.407 & 0.220 \\
HippoRAG2  & 0.256 & 0.196 & \textbf{0.427} & 0.265 & \textbf{0.591} & 0.323 \\
LightRAG   & 0.156 & 0.163 & 0.260 & 0.204 & 0.385 & 0.245 \\
\textsc{LExGraph}       & \textbf{0.292} & \textbf{0.257} & 0.390 & \textbf{0.300} & 0.550 & \textbf{0.354} \\

\bottomrule
\end{tabular}
\end{adjustbox}
\caption{
Evidence retrieval results on our proposed benchmark.
The best result is in bold.
}
\label{tab:retrieval_results}
\end{table}

\section{Details of Experiments}
\subsection{Details of Baselines}
\label{sec:appendix_details_of_baselines}
This section provides details about the baseline methods used in our experiments.
We keep each method's own pipeline and default hyperparameters.
For methods that split the dialogue into fixed passages, each passage is a window of 10 consecutive dialogue turns.
\begin{itemize}[leftmargin=*]
    \item \textbf{ComoRAG} \citep{10.1609/aaai.v40i39.40644} organizes knowledge into veridical, semantic, and episodic layers, iteratively querying them via a metacognitive loop to aggregate cross-layer evidence. We cap the loop at 5 iterations with a 2,000-token per-layer budget.

    \item \textbf{GraphRAG} \citep{edge2025localglobalgraphrag} constructs an entity--relation graph from chunks and clusters them via Leiden to generate hierarchical community summaries.

    \item \textbf{HippoRAG} \citep{NEURIPS2024_6ddc001d} extracts open information extraction triples into a graph, links query entities to graph nodes, and scores passages via Personalized PageRank (PPR; \citealp{10.1145/511446.511513}). Following official defaults, we set the synonym edge threshold to 0.8 and damping factor to 0.5, retrieving the top-5 ranked passages as $X$.

    \item \textbf{HippoRAG2} \citep{3780338.3781176} embeds passage nodes directly into the knowledge graph, links queries to triples, and prunes irrelevant triples with an LLM filter prior to PPR. We retain default settings and retrieve the top-5 passages as $X$.

    \item \textbf{LightRAG} \citep{guo-etal-2025-lightrag} indexes entity--relation graphs and text chunks for dual-level retrieval. We adopt the \textit{mix} mode, retrieving the top-40 graph elements and top-20 chunks as $X$.

    \item \textbf{LightMem} \citep{ICLR2026_a05b7265} manages sensory, short-term, and long-term memory through LLMLingua-2~\citep{pan-etal-2024-llmlingua} compression and offline consolidation. We set compression rate to 0.6 and update threshold to 0.9. We extract memories across both user and assistant turns and retrieve the top-60 entries as $X$.

    \item \textbf{Mem0} \citep{DBLP:conf/ecai/ChhikaraKASY25} dynamically extracts atomic facts from user--assistant turns and resolves them against a persistent memory store. Candidate memories are ranked via hybrid dense--sparse retrieval and entity matching, with the top-200 memories returned as $X$.

    \item \textbf{StructMem} \citep{xu-etal-2026-structmem} organizes dialogue history into an event graph capturing causal and temporal relations. At inference, it activates query-relevant event subgraphs and returns the top-60 event-grounded memory entries as $X$.
\end{itemize}

\subsection{Results on Evidence Retrieval}
\label{sec:appendix_retrieval}
We further examine whether our method and retrieval-based approaches can recover the evidence turns $\mathcal{E}$ annotated in Appendix~\ref{sec:appendix_evidence_annotation}.
Specifically, for each approach, we take its top-$K$ retrieved results ($K \in \{5, 10, 20\}$), each corresponding to a window of 10 consecutive dialogue turns.
For our method, since memory nodes vary in length, each retrieved memory node is mapped to the 10-turn window that contains the node's first dialogue turn, so that all approaches are compared at the same granularity.
A retrieved window is counted as a hit if it contains at least one evidence turn in $\mathcal{E}$.
We report two metrics: (1) Recall@$K$, the proportion of evidence turns in $\mathcal{E}$ that appear in the top-$K$ windows, and (2) NDCG@$K$, which further accounts for the ranks of hit windows.

As shown in Table~\ref{tab:retrieval_results}, our method achieves the highest NDCG under all cutoffs, together with the highest Recall@5.
Meanwhile, HippoRAG2 attains higher Recall at larger cutoffs (e.g., 0.591 vs.\ 0.550 at $K=20$), demonstrating its ability to retrieve evidence scattered across the dialogue.
Nevertheless, our method maintains a higher NDCG even at $K=20$ (0.354 vs.\ 0.323).
Overall, these results demonstrate the strength of our method in precisely locating evidence.

\subsection{Details of Benchmark Analysis}
\label{sec:appendix_benchmark_capability}
We categorize each question into one of six core capabilities based on the primary reasoning process required:
\begin{itemize}[leftmargin=*]
\item \textbf{Fact Retrieval (FR):} Locating an explicit, isolated fact stated directly within conversational turns without requiring multi-step derivation.
\item \textbf{State Tracking (ST):} Monitoring temporal modifications to an entity, parameter, or environment status across turns to identify its latest state.
\item \textbf{Entity Discrimination (ED):} Distinguishing the target entity, attempt, or artifact from structurally or semantically similar candidates accumulated during exploration.
\item \textbf{Numeric Reasoning (NR):} Performing arithmetic calculations, quantitative metric comparisons, or statistical summaries that are not directly stated in the dialogue context.
\item \textbf{Relation Tracing (RT):} Traversing multi-hop associative, hierarchical, or causal links connecting distinct entities or exploration steps.
\item \textbf{Information Aggregation (IA):} Synthesizing multiple pieces of non-contiguous evidence dispersed across prolonged dialogue contexts into a coherent conclusion.
\end{itemize}
Specifically, we employ \texttt{Claude-5.5 Opus}~\citep{anthropic_opus55} to determine the capability most central to answering each question in \textsc{LExMem}.
For each instance, the model is provided with the question $q$, the ground-truth answer $\mathcal{A}$, and the raw dialogue turns corresponding to the golden evidence nodes in $G^*$.
The model is instructed to assign the single capability most central to answering the question.
To verify taxonomy completeness, an \textit{Other} option is included in the candidate set but is never selected across the benchmark, confirming that the six capabilities fully encompass the reasoning requirements in \textsc{LExMem}.

\begin{table}[t]
  \centering
  \resizebox{0.7\columnwidth}{!}{%
  \begin{tabular}{lcccccccc}
    \toprule
    \multirow{2}{*}{\textbf{Setting}} & \multicolumn{2}{c}{\textbf{Dense}} & \multicolumn{2}{c}{\textbf{Sparse}} & \multicolumn{2}{c}{\textbf{Same-Type}} & \multicolumn{2}{c}{\textbf{Overall}} \\
    \cmidrule(lr){2-3}\cmidrule(lr){4-5}\cmidrule(lr){6-7}\cmidrule(lr){8-9}
     & F1 & ACC & F1 & ACC & F1 & ACC & F1 & ACC \\
    \midrule
    Question Only  & 0.104 & 1.31  & 0.088 & 1.16  & 0.082 & 2.69  & 0.090 & 1.73  \\
    Evidence Turns & 0.797 & 78.30 & 0.792 & 78.60 & 0.771 & 60.09 & 0.786 & 72.11 \\
    \midrule
    \textsc{LExGraph} & 0.658 & 70.89 & 0.665 & 66.30 & 0.676 & 55.73 & 0.667 & 63.92 \\
    \bottomrule
  \end{tabular}}
  \caption{Oracle analysis on \textsc{LExMem} using \texttt{Qwen-3.8 27B}. \textit{Question Only} provides only the question, and \textit{Evidence Turns} pairs the question with annotated evidence turns.}
  \label{tab:oracle}
\end{table}

\subsection{Oracle Analysis}
\label{sec:appendix_oracle}
To estimate performance bounds on \textsc{LExMem}, we evaluate two retrieval-free oracle settings using \texttt{Qwen-3.8 27B}:
(1) \textit{Question Only}, which provides solely the question $q$; and
(2) \textit{Evidence Turns}, which pairs $q$ with the gold evidence turns $\mathcal{E}$ (Appendix~\ref{sec:appendix_evidence_annotation}).

As shown in Table~\ref{tab:oracle}, \textit{Question Only} yields near-zero accuracy across all subsets, confirming that \textsc{LExMem} cannot be answered via guesswork or pretrained memorization.
With access to the gold evidence turns $\mathcal{E}$, the model establishes an upper bound, which \textsc{LExGraph} approaches without ground-truth annotations.
Notably, even under \textit{Evidence Turns}, accuracy on \textit{Same-Type} remains substantially lower than on other subsets.
This indicates that the difficulty of \textit{Same-Type} stems not only from locating scattered evidence, but also from the intrinsic challenge of distinguishing and synthesizing similar exploration steps during reasoning.

\begin{table}[t]
  \centering
  \resizebox{0.8\columnwidth}{!}{%
  \begin{tabular}{lcccccccc}
    \toprule
    \multirow{2}{*}{\textbf{Method}} & \multicolumn{2}{c}{\textbf{Dense}} & \multicolumn{2}{c}{\textbf{Sparse}} & \multicolumn{2}{c}{\textbf{Same-Type}} & \multicolumn{2}{c}{\textbf{Overall}} \\
    \cmidrule(lr){2-3}\cmidrule(lr){4-5}\cmidrule(lr){6-7}\cmidrule(lr){8-9}
     & F1 & ACC & F1 & ACC & F1 & ACC & F1 & ACC \\
    \midrule
    \textsc{LExGraph} & 0.658 & 70.89 & 0.665 & 66.30 & 0.676 & 55.73 & 0.667 & 63.92 \\
    \quad w/o Cross-Branch Edges & 0.630 & 67.26 & 0.655 & 65.60 & 0.647 & 49.62 & 0.645 & 60.53 \\
    \quad w/o Neighbors & 0.621 & 67.59 & 0.655 & 65.22 & 0.637 & 47.00 & 0.639 & 59.58 \\
    \bottomrule
  \end{tabular}}
  \caption{Ablation study on \textsc{LExMem} using \texttt{Qwen-3.8 27B}.}
  \label{tab:ablation}
\end{table}

\subsection{Ablation Study}
\label{sec:appendix_ablation}
To evaluate the contribution of graph connectivity in \textsc{LExGraph}, we compare the full framework with two variants on \textsc{LExMem} using \texttt{Qwen-3.8 27B}:
(1) \textit{w/o Cross-Branch Edges}, which removes Type~\uppercase\expandafter{\romannumeral2} edges during graph construction (\S\ref{sec:memory_graph_construction}) and retains only hierarchical edges along the right frontier $\mathcal{F}$; and
(2) \textit{w/o Neighbors}, which does not collect neighboring nodes during retrieval (i.e., $B_{\mathrm{nbr}} = 0$; \S\ref{sec:memory_retrieval}), constructing the context $X$ solely from the primary nodes $\mathcal{M}_{\mathrm{pri}}$ and the edges connecting them.

As shown in Table~\ref{tab:ablation}, removing either component degrades F1 and accuracy across all subsets, and \textit{w/o Neighbors} causes a relatively larger drop than \textit{w/o Cross-Branch Edges}, confirming that both hierarchical and cross-branch edges contribute to answering.
Across question types, the improvements from graph connectivity are concentrated on \textit{Dense} and \textit{Same-Type}: cross-branch edges account for most of the gain on both subsets by linking related steps across separate branches, while hierarchical neighbors provide additional gains on \textit{Same-Type} by connecting sequential steps within a branch.
In contrast, performance on \textit{Sparse} varies only slightly across variants, as its target nodes are separated by multi-hop paths rather than direct neighbors, leaving little for one-hop neighbor collection to recover beyond primary retrieval $\mathcal{M}_{\mathrm{pri}}$.

\subsection{Further Implementation Details}
\label{sec:appendix_implementation_details}
For benchmark curation, we set the mutation probability $p = 0.25$ (\S\ref{sec:evolution_graph}) and the weight of the Jaccard penalty to 0.65 (\S\ref{sec:qa_instance_generation}).
Dialogue lengths after expansion (\S\ref{sec:dialogue_synthesis}) are measured with the \texttt{Qwen-3.8} tokenizer~\citep{qwen38}.
For implementation details of \textsc{LExGraph}, we set $n = 20$, $\tau_{\mathrm{sim}} = 0.75$, $\kappa = 10$, $B_{\mathrm{pri}} = 4096$, and $B_{\mathrm{nbr}} = 4096$.
For evaluation, the ACC judge is GPT-4o mini of version 2024-07-18.

\section{Detailed Prompts}
\label{sec:appendix_prompts}

\begin{figure}[tp]
\centering
\begin{promptbox}{Prompt for Initial Graph Generation (\S\ref*{sec:evolution_graph})}

Generate one empty, structurally plausible graph skeleton for later screening and content instantiation.

\psec{INPUT}
\ph{input\_json}

\psec{REQUIREMENTS}
\begin{pitem}
\item The input contains one empty goal and 3--5 empty proposals. Preserve all supplied IDs and types, and use exactly these goal and proposal nodes. Expand their branches together into one complete graph; do not add or remove goals or proposals.
\item Generate only node types, structure, relations, and traversal order. Do not add concrete content, descriptions, status fields, state-history fields, dialogue, or QA.
\item Encourage many branches, deeper branching, and several long paths. Vary branch lengths and node-type sequences while keeping connections plausible. Avoid repeating one template or adding arbitrary nodes merely to increase size. Choose the total node count freely.
\item An example exploration path is goal $\rightarrow$ proposal $\rightarrow$ hypothesis $\rightarrow$ activity $\rightarrow$ evidence $\rightarrow$ decision. This is one possible branch, not a required template or a sequence of directed semantic relations.
\end{pitem}

\psec{NODE TYPES}
\ph{node\_types}

\psec{RELATIONS (source $\rightarrow$ target)}
\ph{relations}

Consider the proposals jointly. Allow shared prerequisites, activities evaluating multiple proposals, evidence bearing on different branches, and decisions drawing on evidence across proposals where structurally plausible. Add cross-proposal relations when justified, without forcing every pair to connect or using depend\_on as a generic association. Do not force relation coverage or label every adjacent pair in an exploration path.

\psec{TRAVERSAL}
Use \texttt{exploration\_parent\_id} to define one exploration tree rooted at g1; this is ordering metadata, not a semantic relation. The root's parent is null and every other node has one existing parent. All supplied proposals are direct children of the goal, in input order; p1 is the first child. Explore each proposal's subtree before moving to the next proposal. Shared nodes need only one exploration parent; semantic edges can connect them to other branches.

Node order is also the chronological order of the generated dialogue.

Order nodes by left-to-right depth-first preorder: descend through each branch, then backtrack to explore remaining siblings. Sibling order follows appearance in the nodes array. Assign contiguous order values from 1 to N (g1 = 1, p1 = 2); each subtree occupies a contiguous interval. Semantic edges may point backward or across branches without changing this order or duplicating nodes.

\psec{OUTPUT}
Return only one complete JSON object:
\begin{pitem}
\item nodes: objects with exactly id, type, order, \texttt{exploration\_parent\_id}.
\item edges: objects with exactly source, relation, target.
\end{pitem}
Use unique neutral IDs and valid references. Check node counts, seed preservation, relation directions, and depth-first order. No self-edges, duplicate edges, circular prerequisites, Markdown fences, explanations, or omitted sections.
\end{promptbox}
\end{figure}

\begin{figure}[tp]
\centering
\begin{promptbox}{Prompt for Graph Evolution (\S\ref*{sec:evolution_graph})}

Continue this content-free graph with the selected development and return the complete graph, including its unchanged history.

Original seeds (one goal and 3--5 proposals): \ph{seed\_input\_json}

Immutable prefix through order \ph{cutoff\_order}: \ph{prefix\_json}

\psec{Requirements:}
\begin{pitem}
\item Preserve every prefix node and edge. Add no edges between two prefix nodes. Regenerate the future; the discarded suffix need not be reconstructed.
\item Include exactly the supplied goal and proposals, including those not yet in the prefix. Add no descriptions, domain text, status/history fields, dialogue, QA, or extra metadata.
\item Encourage multiple branches, depth, and broad exploration without padding. Place the requested development naturally, possibly after other exploration, then continue its consequences and complete the remaining graph.
\item Templates specify outcomes, not fixed node chains. Choose the intervening structure, node count, and connections freely within the schema.
\item New nodes have unique neutral IDs and consecutive orders after the cutoff. Order is dialogue chronology. Activities precede their new outputs; evaluated targets precede their new evaluating activities. Revisions/revivals follow their targets.
\item Keep a single exploration tree rooted at g1 and left-to-right DFS preorder. The root has a null parent; every other node has one earlier existing parent. Proposals are direct children of g1 in seed order, each linked from g1 by decompose. Continue from the traversal frontier without reopening closed subtrees; use semantic edges to refer to earlier topics across branches. Exploration parents specify traversal only, not semantic relationships.
\item Preserve old findings. New evidence does not automatically invalidate related nodes or settle conflicting evidence. Account for explicit prerequisites without silently rewriting historical dependencies.
\item revise does not imply adoption or abandonment. refute alone does not justify revive, which requires a previously rejected, deferred, or inactive target. Evidence supporting a decision does not specify what that decision selected.
\end{pitem}

\psec{Node types:}
\ph{node\_types}

\psec{Allowed relations (source $\rightarrow$ target):}
\ph{relations}

Return only JSON with nodes and edges. Each node has exactly id, type, order, \texttt{exploration\_parent\_id}; each edge has exactly source, relation, target. Check prefix preservation, seed coverage, valid references and relation directions, DFS/chronological order, and completion of the selected task. No self-edges, duplicate edges, circular prerequisites, Markdown fences, or explanations.

\psec{SELECTED TASK}
\ph{mutation\_rule}
\end{promptbox}
\end{figure}

\begin{figure}[tp]
\centering
\begin{promptbox}{Prompt for Node Instantiation (\S\ref*{sec:dialogue_synthesis})}

Fill the current node with concrete content for the supplied task.

\psec{INPUT}
\texttt{[task]}: \ph{task}\\
\texttt{[prefix\_graph]}: \ph{prefix\_graph}\\
\texttt{[current\_node]}: \ph{current\_node}\\
\texttt{[current\_edges]}: \ph{current\_edges}\\
\texttt{[retry\_feedback]}: \ph{retry\_feedback}

\psec{CONTENT}
Write one English sentence or phrase, at most 20 words. State the specific information this node adds:
\ph{node\_type\_guidelines}

\psec{CONCRETE DETAILS}
Actively invent plausible task-specific details: metric names, numerical values, experiment or dataset names, methods, and versions. For measurable claims and findings, prefer a named metric with a value, threshold, or comparison over vague statements such as ``performs better.'' For activities and artifacts, use concrete test, dataset, or deliverable names when they help identify the work.

Example: ``On CiteBench, caption links raised evidence recall from 69\% to 84\%.''

Choose details that help express the current node or its relations; not every node needs a number or new name. Reuse established names and metrics across related nodes. Keep invented measurements plausible, units clear, and recorded facts consistent. These details are fictional, not reports of real experiments. The 20-word limit still applies.

Use earlier content as context. Name the relevant method or object briefly; do not repeat its full description, the task requirements, or previous results. Include a condition, threshold, or earlier value only when needed to distinguish the current claim or comparison. Do not turn the same goal statement into successive questions, hypotheses, and plans by changing its wording.

\psec{RELATIONS (source $\rightarrow$ target)}
Satisfy every current edge using the current content and prefix together:
\ph{relations}

The exploration parent specifies traversal, not a semantic relation. Do not invent earlier events or anticipate future outcomes. New findings may challenge earlier conclusions without rewriting earlier observations. On retry, change only the current content to address feedback. Do not add graph fields, keywords, or dialogue.

\psec{OUTPUT}
Return only JSON: \texttt{\{"node\_id": "<current node id>", "content": "<sentence or phrase>"\}}
\end{promptbox}
\end{figure}

\begin{figure}[tp]
\centering
\begin{promptbox}[\footnotesize]{Prompt for Dialogue Expansion}

Expand the current node into an English user--assistant dialogue fragment.

\psec{INPUT}
\texttt{[previous\_dialogue]}: \ph{previous\_dialogue}\\
\texttt{[current\_node]}: \ph{current\_node}\\
\texttt{[earlier\_relations]}: \ph{earlier\_relations}\\
\texttt{[continuation]}: \ph{continuation}\\
\texttt{[retry\_feedback]}: \ph{retry\_feedback}

\psec{CONTENT}
Express the current node fully and accurately. Preserve its specific names, numerical values, units, conditions, and any stated comparisons or corrections, without replacing them with vague summaries. Natural paraphrasing is allowed; preserve meaning and certainty: hypotheses remain hypotheses, and plans are not completed experiments.

Aim for a long, rich, multi-turn exchange centered on the current node, and extend it as far as you naturally can: the longer and richer the discussion, the better, as long as every turn stays faithful to the current node and the rules below and adds no new task-specific facts. Freely develop its meaning through natural discussion. Verbosity, repeated explanations, paraphrases, and clarification are acceptable when faithful to the supplied information. There is no fixed message count.

The assistant may answer at length, in several sentences or short paragraphs, for example by explaining what a term or condition means, what the node does and does not say, how a value or threshold can be read, or by walking through a small general illustration. The user can ask short follow-up questions, request clarification, restate the idea in their own words, or probe a detail, and the discussion can continue through several such rounds. These are only suggestions; develop the node in whatever natural way fits, as long as the supplied information stays accurate.

Stay within the current node and established history. General conceptual discussion is allowed, but do not invent or change task-specific facts, experiments, data, causes, implementation details, commitments, or decisions, or advance to an unsupported next step.

\psec{HISTORICAL INFORMATION}
Earlier dialogue and relations provide context for consistency; they are not additional content to narrate. Apply these rules to both user and assistant:
\begin{pitem}
\item Do not declare an earlier result, hypothesis, plan, or branch valid, invalid, supported, refuted, obsolete, superseded, abandoned, or revived unless that specific judgment is explicitly in \texttt{current\_node}.
\item Do not tell the reader which historical value or version is now authoritative, which old value must not be used, or which earlier target is now met or missed unless \texttt{current\_node} explicitly states that conclusion.
\item Do not reconstruct a history of changes or provide a recap of earlier results to justify the current node. This also applies to decision nodes and closing summaries.
\item Do not generate questions that invite any of the above. A fact's presence in \texttt{previous\_dialogue} or a relation's logical implication does not authorize restating it as a historical status judgment here.
\end{pitem}
Preserve every comparison, correction, decision, and reason explicitly in \texttt{current\_node}, but do not extend its scope. A node that states a correction may express that correction; later nodes must not retell it unless they also explicitly contain it. A phrase such as ``after corrected audits'' does not authorize reconstructing those audits and their old values.

Historical references may identify the method, experiment, object, or baseline needed to understand the current node. Give only the necessary reference; do not add which older alternative was superseded or why it is no longer valid. For example, identifying a comparison baseline does not authorize saying ``use this value instead of the obsolete earlier value.'' Do not expose graph metadata.

\psec{CONTINUATION}
Write only the new fragment. Begin with \texttt{continuation.next\_role} and alternate user and assistant. Either role may end the fragment. Continue naturally across node and session boundaries without forced greetings, closing summaries, or complete question--answer pairs. Avoid questions requiring unavailable future information.

\psec{OUTPUT}
Return only JSON: \texttt{\{"messages": [\{"role": "user", "content": "..."\}, \{"role": "assistant", "content": "..."\}]\}}

The roles above are illustrative; follow \texttt{continuation.next\_role}. The program assigns source-node labels; do not generate them.
\end{promptbox}
\end{figure}

\begin{figure}[tp]
\centering
\begin{promptbox}{Prompt for QA Instance Generation (\S\ref*{sec:qa_instance_generation})}

Generate exactly one question about \ph{target}.

\ph{input\_json}

\psec{Dense subgraphs}
subgraph contains the sampled nodes and their original directed relations. Consider drawing on information from multiple nodes and the relations among them when forming the question. Use \texttt{full\_graph} to check relevant context and later updates; do not treat the subgraph as the complete history. Edge density and \texttt{selection\_signals} are sampling hints, not proof of causality or difficulty.

\psec{Sparse subgraphs}
subgraph contains a sampled path whose endpoints connect through the goal. \texttt{path\_node\_ids} lists this path; \texttt{endpoint\_ids} identifies its endpoints. The path was sampled without edge directions: connectivity alone does not establish a causal chain. You are encouraged to connect information from both sides of the goal and draw on relevant context, revisions, or dependencies from intermediate nodes.

\psec{Same-type sets}
\texttt{focus\_nodes} contains all nodes of \texttt{target\_type}. Prefer questions where the answer depends on distinguishing among several same-type nodes, especially when they represent alternatives, revisions, competing measurements, condition changes, or later corrections. Use \texttt{full\_graph} to check context and later updates; avoid questions already answered by a single node.

\psec{Requirements}
\begin{pitem}
\item Ask one clear question in English, ending with exactly one question mark. Ask only one substantive question: do not combine multiple requests or stack subquestions, even under a single question mark. Do not reveal the answer or mention node IDs in the question. Avoid asking the respondent to explain, elaborate on, or justify an answer.
\item Include the context, conditions, and stage of the dialogue needed to establish the question's scope without revealing the answer.
\item Keep the reference answer aligned with what the question asks. Include the requested information and the units and qualifiers needed to interpret it. Avoid silently adding background facts or omitting requested details.
\item Express each answer item as a short entity, value, status, or phrase. Separate independently scorable facts without expanding them into explanations.
\item Use \texttt{named\_slots} for distinct fields or \texttt{unordered\_set} for a list of items. \texttt{answer\_keywords} must use the same keys as answer; canonical must equal the corresponding answer, and aliases must be equivalent expressions.
\item Cite supporting node IDs for each answer item. The question must be supported by the graph. Return exactly one item in questions, or an empty array if no supported question is possible.
\end{pitem}

Return JSON only:\\
\texttt{\{"questions": [\{"question": "... ?", "answer\_mode": "named\_slots or unordered\_set", "answer": \{"slot\_name": "short value"\}, "answer\_keywords": \{"slot\_name": \{"canonical": "short value", "aliases": []\}\}, "evidence\_node\_ids": \{"slot\_name": ["node ID"]\}\}]\}}
\end{promptbox}
\end{figure}

\begin{figure}[tp]
\centering
\begin{promptbox}{Prompt for Dialogue Segmentation (\S\ref*{sec:memory_graph_construction})}

Decide where the newly read turns start new memory units in an ongoing conversation.

\ph{input\_json}

\texttt{open\_unit\_turns} is the complete unit already open before this batch (null if this is the first batch). \texttt{new\_turns} are the turns just read. Each turn has a \texttt{turn\_number} and the user's message of that turn; the assistant's replies are omitted, but each turn is a user message plus a reply, so a short user follow-up still belongs to the discussion before it.

Go through \texttt{new\_turns} in order and judge only whether the discussion focus shifts --- a change in the topic, task, question, or piece of work being discussed, not a change in tone, a follow-up question, or a brief detour that returns to the same focus. Do not use turn count, message length, or role alternation as a signal. Judge each hypothesis, experiment, or measurement round on its own: a new hypothesis being proposed, a new experiment being run, or a new result being reported is a new piece of work and starts a new unit, even if it concerns the same broad subject as \texttt{open\_unit\_turns} or an earlier round in this batch.
\begin{pitem}
\item \texttt{new\_unit\_starts}: \texttt{turn\_number} values, from \texttt{new\_turns} only, where a new unit begins. Each one closes the unit open immediately before it (\texttt{open\_unit\_turns} for the first split point in this batch, otherwise the segment right before it). May be empty, or list several if the focus shifts more than once in this batch.
\end{pitem}
Return JSON only: \texttt{\{"new\_unit\_starts": [turn\_number]\}}
\end{promptbox}
\end{figure}

\begin{figure}[tp]
\centering
\begin{promptbox}{Prompt for Key Generation (\S\ref*{sec:memory_graph_construction})}

Write the Key for a memory unit that has just closed in an ongoing conversation.

\ph{input\_json}

messages lists this unit's turns in order.

Produce a title, summary, and key facts that let this unit be recognized and searched for later without rereading its original text.
\begin{pitem}
\item title: one line, a few words naming what this unit is about.
\item summary: 1--2 sentences covering what was discussed and, if applicable, what was concluded or decided.
\item \texttt{key\_facts}: short, independent statements. Preserve every specific number, name, identifier, condition, and conclusion stated in the unit; do not round, generalize, or paraphrase them away. Use only the supplied messages.
\end{pitem}
Return JSON only: \texttt{\{"title": "...", "summary": "...", "key\_facts": ["..."]\}}
\end{promptbox}
\end{figure}

\begin{figure}[tp]
\centering
\begin{promptbox}{Prompt for Edge Construction (\S\ref*{sec:memory_graph_construction})}

Decide which earlier memory units connect to a newly closed unit, and pick its parent.

\ph{input\_json}

\texttt{new\_unit} gives the closed unit's Key (title, summary, \texttt{key\_facts}). candidates lists earlier units, each with \texttt{unit\_id}, title, summary, and \texttt{on\_right\_frontier}; right-frontier candidates are listed first, in order from earliest to most recent, representing the path the conversation has been building on. The remaining candidates were retrieved by similarity and are not on the right frontier --- units the conversation has since moved past but that may still be directly relevant.

For every candidate directly related to \texttt{new\_unit}, report:
\begin{pitem}
\item \texttt{unit\_id}, exactly as given in candidates.
\item direction: whether the relation reads from the new unit to the candidate, or the reverse. This follows the semantic relationship, not chronological order --- a later unit can point back at an earlier one and an earlier unit can point forward at a later one.
\item relation: one sentence phrased in the chosen direction, as ``<source> ... <target>''.
\end{pitem}

\psec{Requirements:}
\begin{pitem}
\item Include at least the parent among the connected candidates.
\item Only use \texttt{unit\_id} values from candidates; never invent one or connect the new unit to itself.
\item \texttt{parent\_id}: exactly one candidate with \texttt{on\_right\_frontier=true}, the item this new unit continues from, and it must also appear among the connected candidates with an edge to the new unit. This is a hard requirement even when a similarity-retrieved candidate (\texttt{on\_right\_frontier=false}) looks more topically related: if that is the case, still pick the right-frontier item this new unit most directly continues from as \texttt{parent\_id}, and separately add an edge to the more topically-related non-frontier candidate to capture that relationship.
\end{pitem}

Return JSON only:\\
\texttt{\{"edges": [\{"unit\_id": "candidate unit ID", "direction": "new\_to\_candidate or candidate\_to\_new", "relation": "one sentence phrased in the chosen direction"\}], "parent\_id": "unit ID of the chosen parent"\}}
\end{promptbox}
\end{figure}

\begin{figure}[tp]
\centering
\begin{promptbox}{Prompt for ACC Evaluation (\S\ref*{sec:evaluation_metrics})}

I will give you a question, the correct answer, and a response from a model. Grade the response as yes (fully correct), partial (partly correct) or no (wrong).

The correct answer is split into required parts, each with acceptable aliases. A part counts as stated if the response gives it with the same meaning and its specific details (names, quantities, conditions); paraphrases, synonyms and the listed aliases are acceptable, details already given in the question may be omitted, and numbers must match exactly. If the question already states the unit (for example, it asks how many percentage points), a bare number counts as stating it.
\begin{pitem}
\item yes: the response states every required part, and contains nothing wrong or irrelevant.
\item partial: the response states some of the required parts (or only part of them) but misses the rest, and contains nothing wrong or irrelevant.
\item no: anything else, including a response that states none of the parts, contradicts a part, hedges between alternative answers, or contains any wrong or irrelevant content (information the question does not ask for, such as extra items, other findings or unrelated details), even if it also states the correct answer.
\end{pitem}
Check before you grade. First, for each required part (using the given part ids), say whether the response states it: yes, partly or no. Then list every piece of the response that is wrong or irrelevant (an empty list if there is none). Finally, give the grade that follows from these checks.

Return only JSON: \texttt{\{"checks": [\{"id": "part id", "stated": "yes, partly or no"\}], "wrong\_or\_irrelevant": ["..."], "correct": "yes, partial or no"\}}

\texttt{[question]}: \ph{question}

\texttt{[correct\_answer\_parts]}: \ph{answer\_parts\_with\_aliases}

\texttt{[model\_response]}: \ph{answer\_items}
\end{promptbox}
\end{figure}

\end{document}